\documentclass[lettersize,journal]{IEEEtran}
\usepackage[utf8]{inputenc}
\usepackage{amsmath,amsfonts}
\usepackage{algorithmic}
\usepackage{array}
\usepackage{cite}
\usepackage{tabularx}
\usepackage{booktabs}
\usepackage{booktabs}
\usepackage{multirow}
\usepackage{tabularx}
\usepackage{graphicx}              
\usepackage{subcaption}            
\usepackage{float}                 
\usepackage[caption=false,font=normalsize,labelfont=sf,textfont=sf]{subfig}
\usepackage{textcomp}
\usepackage{stfloats}
\usepackage{url}
\usepackage{verbatim}
\usepackage{graphicx}

\def\BibTeX{{\rm B\kern-.05em{\sc i\kern-.025em b}\kern-.08em
    T\kern-.1667em\lower.7ex\hbox{E}\kern-.125emX}}
\usepackage{balance}
\begin{document}
\title{Adversarial Patch Attack for Ship Detection via Localized Augmentation}
\author{Chun Liu, Panpan Ding, Zheng Zheng, Hailong Wang, Bingqian Zhu, Tao Xu, Zhigang Han, Jiayao Wang
\thanks{Chun Liu is with the State Key Laboratory of Spatial Datum,  College of Remote Sensing and Geoinformatics Engineering, Faculty of Geographical Science and Engineering, Henan University, Zhengzhou 450046, China, and the National Key Laboratory of Integrated Technology on Aircraft Control, Beihang University, Beijing 100191, China. Panpan Ding, Hailong Wang, and Bingqian Zhu are with the School of Computer and Information Engineering, Henan University, Kaifeng 475004, China. Zheng Zheng is with the National Key Laboratory of Integrated Technology on Aircraft Control,School of Automation Science and Electrical Engineering, Beihang University, Beijing 100191, China. Tao Xu is with the State Key Laboratory of Spatial Datum, Henan University, Zhengzhou 450046, China, and the School of Computer and Information Engineering, Henan University, Kaifeng 475004, China. Zhigang Han and Jiayao Wang are with the State Key Laboratory of Spatial Datum, College of Remote Sensing and  Geoinformatics Engineering, Faculty of Geographical Science and Engineering, Henan University, Zhengzhou 450046, China.
(e-mail:liuchun@henu.edu.cn;1823373789@qq.com;zhengz@buaa.edu.cn; whl0032@qq.com;104754231411@henu.edu.cn;txu@henu.edu.cn;
zghan@henu.edu.cn; )}}

\markboth{}%
{How to Use the IEEEtran \LaTeX \Templates}
\maketitle

\begin{abstract}
Current ship detection techniques based on remote sensing imagery primarily rely on the object detection capabilities of deep neural networks (DNNs). However, DNNs are vulnerable to adversarial patch attacks, which can lead to misclassification by the detection model or complete evasion of the targets. Numerous studies have demonstrated that data transformation-based methods can improve the transferability of adversarial examples. However, excessive augmentation of image backgrounds or irrelevant regions may introduce unnecessary interference, resulting in false detections of the object detection model. These errors are not caused by the adversarial patches themselves but rather by the over-augmentation of background and non-target areas. This paper proposes a localized augmentation method that applies augmentation only to the target regions, avoiding any influence on non-target areas. By reducing background interference, this approach enables the loss function to focus more directly on the impact of the adversarial patch on the detection model, thereby improving the attack success rate. Experiments conducted on the HRSC2016 dataset demonstrate that the proposed method effectively increases the success rate of adversarial patch attacks and enhances their transferability.
\end{abstract}


\section{Introduction}
\IEEEPARstart With the rapid development of remote sensing technology, the volume of imagery and video collected by Earth observation satellites and aerial platforms has grown significantly. Detecting ship targets from these images is of great significance for sea traffic monitoring, ship search and rescue, fisheries management, and maritime situational awareness~\cite{li2021ship}. 

Deep neural networks (DNNs) now have shown impressive performance for ship detection ~\cite{lei2022groupnet}, providing effective tools for feature extraction from remote sensing images. However, DNNs are vulnerable to adversarial attacks which have raised widespread concern. Adversarial attacks aim to mislead models by introducing small perturbations to inputs \cite{szegedy2013intriguing}. Based on the attacker's knowledge, adversarial attacks can be classified as white-box (full access to the model) or black-box (interaction through input-output only). Since most deployed systems conceal internal structures, black-box attacks present more realistic threats. They often rely on the transferability of adversarial examples—input samples crafted on a surrogate model that can deceive other models. 

Adversarial attacks can be also categorized into pixel-based and patch-based types. Pixel-based attacks modify part or all of an image’s pixels~\cite{ghosh2022blackbox}, often remaining imperceptible to humans~\cite{chow2020adversarial}. Xie et al.\cite{xie2017adversarial} proposed DAG, the first adversarial method for object detection. Li et al.\cite{li2018robust} introduced RAP, which disables region proposal networks to reduce detection accuracy. Wei et al.~\cite{wei2018transferable} utilized GANs to quickly generate examples. Patch-based attacks apply adversarial patches to image regions to induce incorrect predictions. Compared to pixel perturbations, patches are more physically realizable, as they can be printed or applied in real scenes. Brown et al.\cite{brown2017adversarial} introduced adversarial patches targeting classifiers. Liu et al.\cite{liu2018dpatch} proposed DPatch for attacking both localization and classification. Lee and Kolter~\cite{lee2019physical} later improved patch update methods for better suppression of detection outputs.

%To improve the transferability of patch-based attacks, Various data augmentation strategies have been proposed. For example, Xie~\cite{xie2019improving} applied random scaling and padding, Dong et al.\cite{dong2018boosting} used momentum iterations. Guo\cite{guo2020simple} and Lin~\cite{lin2020nesterov} introduced input transformations and low-frequency filtering to enhance robustness.

%Most methods are designed for white-box settings and neglect black-box performance gaps.

However, existing patch-based methods mainly focus on natural images, and less attention has been paid to physical attacks on ships detection in remote sensing imagery. Moreover, to improve the transferability of black-box adversarial attacks, data augmentation strategy has been widely used  \cite{ dong2018boosting, guo2020simple, lin2020nesterov}. But current methods apply the strategy to the whole images. This may disturb background regions excessively and result in false detections of the object detection model. Nevertheless, the errors of such false detections  are not caused by the adversarial patches themselves but rather by the over-augmentation of background and non-target areas,  destabilizing the adversarial patch optimization and weakening attack effectiveness.

\begin{figure}[H]
  \centering
  \begin{subfigure}{0.32\linewidth}
    \includegraphics[width=\linewidth]{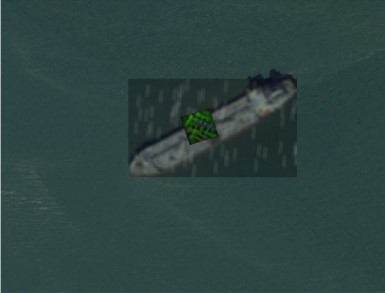}
  \end{subfigure}
  \hfill
  \begin{subfigure}{0.32\linewidth}
    \includegraphics[width=\linewidth]{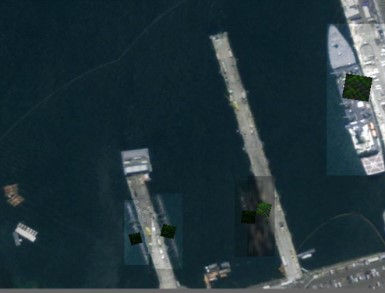}
  \end{subfigure}
  \hfill
  \begin{subfigure}{0.32\linewidth}
    \includegraphics[width=\linewidth]{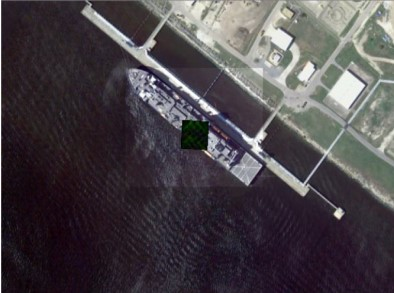}
  \end{subfigure}
  \caption{ Examples of images after localized data augmentation processing}
  \label{fig:1}
\end{figure}

Focusing on the attacks to ship detection from remote sensing images and revealing the vulnerability of the detection models, this paper propose a  adversarial patch generation method via localized augmentation. The  proposed method applies transformations only to target regions, as shown in Fig.1, avoiding any influence on non-target areas. By reducing background interference, this approach enables the loss function to focus more directly on the impact of the adversarial patch on the detection model, thereby improving patch optimization stability and attack performance. Using HRSC2016 dataset, experiments on three YOLOv5 variants (YOLOv5-M, YOLOv5-S, YOLOv5-N) demonstrate that the proposed method significantly improves attack success rates. Transferability experiments show that patches generated via localized augmentation exhibit stronger cross-model attack capability, improving robustness in diverse remote sensing scenarios.

\section{Problem Definition}

Given a clean image \( I \), the attacker aims to optimize a malicious perturbation \( E \), which is added to the original image to produce an adversarial example \( I' \), thereby misleading the object detector to make incorrect predictions. In this paper, the generation of the adversarial perturbation \( E \) (i.e., the adversarial patch) can be conceptually described as:

\begin{equation}
\begin{split}
I' = I \odot (1 &- M) + E \odot M\\
f(I') &\ne f(I) 
\end{split}
\label{eq:1}
\end{equation}

where \( \odot \) denotes element-wise multiplication, \( M \) is the binary mask used to apply the patch, and \( f \) represents the victim model whose parameters remain unchanged during the attack. The generated adversarial example \( I' \) causes the model to make incorrect predictions, i.e., \( f(I') \ne f(I) \).

\begin{figure*}[t] % * 号表示跨双栏，t表示放在页面顶部
  \centering
  \includegraphics[width=0.7\textwidth]{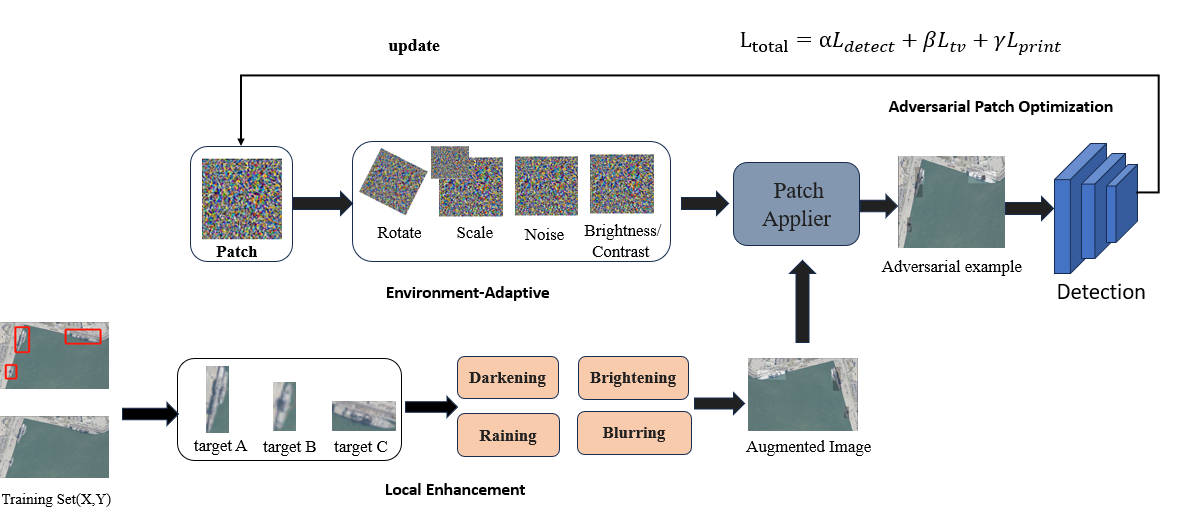} % 图片宽度占满整个双栏宽度
  \caption{Framework for generating adversarial patches based on local data augmentation}
  \label{fig:2}
\end{figure*}

\section{Methodology}
The proposed local augmentation-based adversarial patch generation framework, as shown in Fig.~\ref{fig:2}, consists of three main modules: local data augmentation, environmental adaptation, and adversarial patch optimization. First, training images are processed by the local data augmentation module, where target regions are extracted and randomly transformed using operations such as darkening, brightening, simulated rainfall, and blurring, generating target regions under diverse environmental conditions. These transformed regions are then fused with the original images using a mask matrix to form augmented adversarial samples. Next, the environmental adaptation module adjusts factors such as contrast, brightness, and noise to simulate various physical conditions, thereby improving the robustness of the adversarial patches. Finally, the adversarial patch optimization module refines the initial patch by minimizing non-printability score, total variation loss, and detection loss, producing the final adversarial patch.

\subsection{Local Data Augmentation Module}

This paper proposes a local augmentation-based data processing method to improve the effectiveness and robustness of adversarial patches. Given a training set $\mathcal{D} = \{(x_i, y_i)\}_{i=1}^N$, where $x_i$ is the input image and $y_i$ is the corresponding target label (including bounding box information), the target regions are first extracted based on the annotations. For each image $x$, such extraction process can be described as Eq. (2). 

\begin{equation}
R = \bigcup_{k=1}^{n} r(y_k)
\end{equation}

 where $y_k$ denotes the annotation of the $k$-th object, $r(y_k)$ is the pixel coordinate set of the target region computed from the label, $R$ is the union of all target region pixel coordinates, and $n$ is the total number of target objects in the image.

Then,  one data augmentation operation such as weather change, blurring, or brightness adjustment is randomly selected and applied to each target region $r_j$, as shown in Eq. (3), where $T_j$ denotes the randomly selected augmentation operation. Each object is transformed differently to increase diversity.

\begin{equation}
R_{\text{aug}} = \bigcup_{j=1}^{n} T_j(r_j), \quad r_j \in R
\end{equation}

Finally, the augmented regions are placed back to the original image to form the augmented adversarial samples. The process is described as follows:

\begin{equation}
x' = (1 - M) \odot x + M \odot A(R)
\end{equation}

where $M$ is the mask of the transformed region (1 for target pixels, 0 for background), $\odot$ represents element-wise multiplication, and $A(R)$ is the result of augmenting the original target regions $R$. This local augmentation strategy avoids background interference introduced by global transformations and better focuses the adversarial optimization on the object area, thus improving attack performance and transferability.

\subsection{Environmental Adaptation Module}

Due to varying image quality in maritime remote sensing scenarios, adversarial patches need to adapt to different environments. We introduce a scene intensity matching method by adjusting the contrast, brightness, and noise of the patch to fit the environment characteristics. Based on the printable adversarial patch $P$, we perform scene adaptation as:

\begin{equation}
P = ((P \times \text{scene}_c) + \text{scene}_b) + \text{scene}_n
\end{equation}

where $\text{scene}_c$ is the scene contrast factor, $\text{scene}_b$ is the brightness shift, and $\text{scene}_n$ is the added uniform noise. Adjusting these parameters improves patch effectiveness in varying scenes, enhancing the success rate of attacks.

After intensity matching, we apply affine transformations to the patches, including scaling to fit the target bounding box and rotation to simulate viewing angles of ship images. These transformations enhance the robustness of the patches in real-world physical attacks.

\subsection{Adversarial Patch Optimization}

To optimize the adversarial patches to degrade the performance of ship detection systems in remote sensing imagery, we adopt a combined loss function that includes non-printability score loss $L_{\text{nps}}$,  total variation loss $L_{\text{tv}}$, and detection loss $L_{\text{det}}$.

The \textbf{non-printability score loss} $L_{\text{nps}}$ measures whether the patches can be realistically printed:

\begin{equation}
L_{\text{nps}} = \sum_{p \in P} \min_{\zeta \in \mathcal{Z}_{\text{print}}} \|p - \zeta\|^2
\end{equation}

where $p$ is a pixel in the patch $P$, and $\mathcal{Z}_{\text{print}}$ is the set of printable colors. This loss minimizes the distance between patch pixels and printable colors to ensure fidelity during physical realization.

The \textbf{total variation loss} ensures that the generated adversarial patch maintains visual smoothness, thereby avoiding abrupt pixel transitions and making the patch appear more natural. The total variation Loss is computed as follows:

\begin{equation}
{L}_{\text{TV}} = \sum_{i=1}^{N-1} \sum_{j=1}^{N-1} \left( (P_{i,j} - P_{i+1,j})^2 + (P_{i,j} - P_{i,j+1})^2 \right)
\label{eq:7}
\end{equation}

where \( N \) is the total number of pixels in the patch, and \( P_{i,j} \) denotes the pixel value at position \( (i,j) \).

To effectively reduce the accuracy of the object detector, the \textbf{detection loss} is also included to minimize the maximum confidence score of the target object detected by the model. The loss is defined as:

\begin{equation}
{L}_{\text{tv}} = \max_{b \in B} S(f(I'), b)
\label{eq:8}
\end{equation}

where \( S(f(I'), b) \) denotes the confidence score output by the detection model \( f \) for the bounding box \( b \) on the adversarial example \( I' \), and \( B \) is the set of all detected bounding boxes.

The final optimization objective is a weighted sum:

\begin{equation}
L = \alpha L_{\text{det}} + \beta L_{\text{tv}} + \gamma L_{\text{nps}}
\end{equation}

where $\alpha$, $\beta$, and $\gamma$ are hyperparameters that balance attack effectiveness, visual smoothness, and printability.

\section{EVALUATION}
\subsection{Datasets}
This study conducts experiments on the HRSC2016 dataset, a publicly available high-resolution remote sensing dataset specifically designed for ship detection. HRSC2016 contains remote sensing ship images collected from various scenarios such as ports and open sea areas. The dataset includes 1,061 high-resolution images with a total of 2,976 ship targets. The training, validation, and test sets consist of 436, 181, and 444 images, respectively. The annotations are provided in both horizontal bounding box (HBB) and oriented bounding box (OBB) formats, making it suitable for rotated object detection tasks. Since the image resolutions in HRSC2016 range from 300×300 to 1500×900, we perform data preprocessing before training by removing objects that occupy less than 0.05\% of the image area to ensure a reasonable patch region on each object. The images are resized to 640×640 with gray pixel padding to preserve the aspect ratio.

\subsection{Threat Model}
We trained three YOLOv5-based detectors—YOLOv5-N, YOLOv5-S, and YOLOv5-M—on the HRSC2016 dataset. The YOLOv5~\cite{jocher2022yolov5} series is widely adopted in remote sensing scenarios due to its favorable trade-off between accuracy and computational efficiency. The three models represent small, medium-small, and medium-sized variants, respectively, allowing comprehensive evaluation of patch performance across different model complexity. When measured at an IoU threshold of 0.5, the average precisions are 94.8 (YOLOv5-N), 96.6 (YOLOv5-S), and 96.9 (YOLOv5-M). These results indicate that the detectors trained on the HRSC2016 dataset provide a strong foundation for generating robust adversarial patches.

\subsection{Evaluation Metrics}
To evaluate attack effectiveness, we adopt commonly used metrics including average precision (AP), recall (R), and attack success rate (ASR), following the standard definitions in \cite{liu2018dpatch}.
\subsection{Experimental Setup}
In this experiment, the input size of the model is set to 640×640, and training is conducted using Automatic Mixed Precision (AMP). The initial size of the adversarial patch is set to 64×64, with pixel values constrained between 0 and 255. The patch is initially initialized in grayscale mode and then converted to RGB format. During training, the patch position is randomized, and it undergoes various transformations including rotation and translation, with an offset range of [-0.1, 0.1]. Additionally, the target-patch size ratio is set to 0.12, and the patch position is updated in each training iteration. The optimizer used is Adam, with an initial learning rate of 0.03, and the training process spans 200 epochs. This experiment focuses on single-class object detection, specifically targeting the "ship" category.
\subsection{Comparison Experiments}

\begin{figure}[htbp]
  \centering
  % 第一行三张图
  \begin{minipage}[b]{0.3\linewidth}
    \centering
    \includegraphics[width=\linewidth]{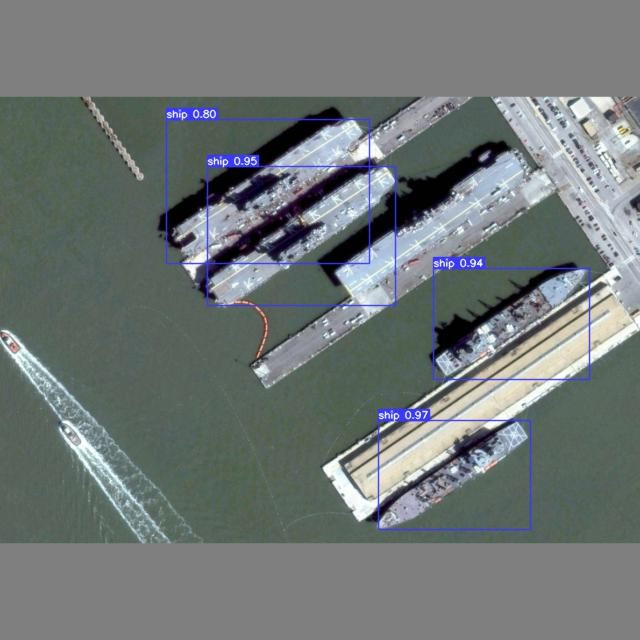}
  \end{minipage}
  \hfill
  \begin{minipage}[b]{0.3\linewidth}
    \centering
    \includegraphics[width=\linewidth]{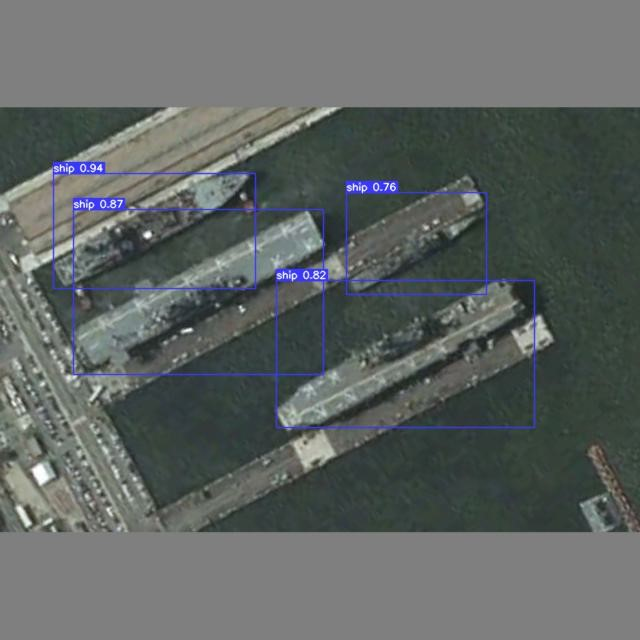}
  \end{minipage}
  \hfill
  \begin{minipage}[b]{0.3\linewidth}
    \centering
    \includegraphics[width=\linewidth]{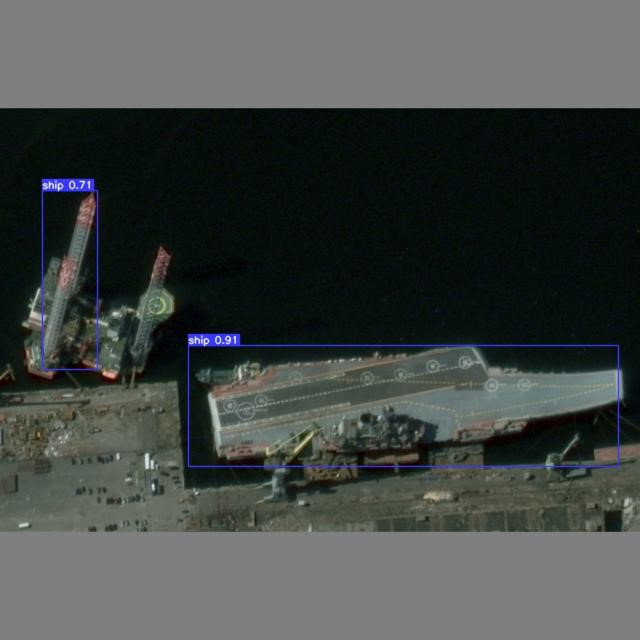}
  \end{minipage}

  \vspace{0.5em}

  % 第二行三张图
  \begin{minipage}[b]{0.3\linewidth}
    \centering
    \includegraphics[width=\linewidth]{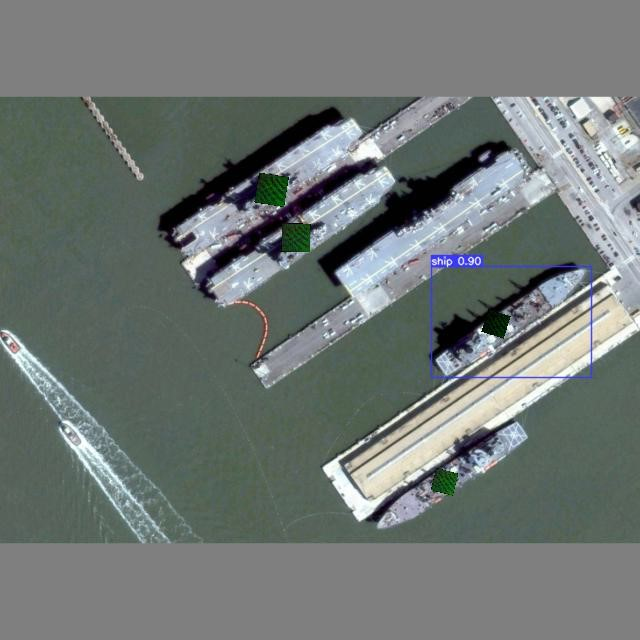}
  \end{minipage}
  \hfill
  \begin{minipage}[b]{0.3\linewidth}
    \centering
    \includegraphics[width=\linewidth]{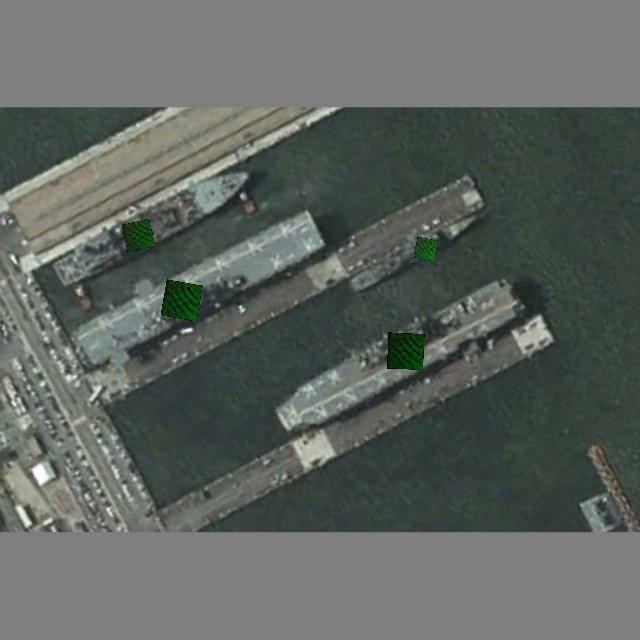}
  \end{minipage}
  \hfill
  \begin{minipage}[b]{0.3\linewidth}
    \centering
    \includegraphics[width=\linewidth]{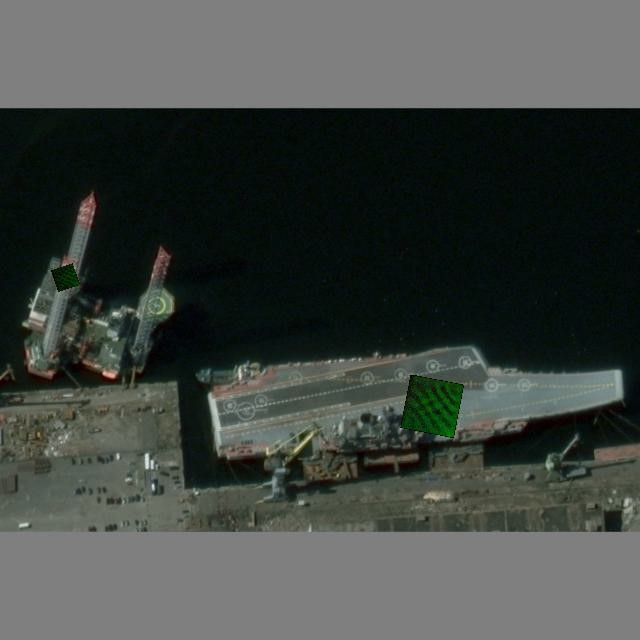}
  \end{minipage}

  \caption{Performance of YOLOv5-N for detecting ships from the images of HRSC2016 dataset. The first row presents the results from clean images, while the second row shows the results after applying the adversarial patches. It can be clearly seen that after adding the adversarial patches, the ship detector exhibits a significant number of missed detections.}
  \label{fig:3}
\end{figure}

We train adversarial patches separately for YOLOv5-M, YOLOv5-S, and YOLOv5-N, and evaluate their impact on recall, average precision, and attack success rate (ASR) under white-box settings. To assess whether performance degradation stems from adversarial characteristics rather than mere occlusion, we compare the effectiveness of our patches with that of random noise patches. Fig.~\ref{fig:3} illustrates the attack effects of our patches on HRSC2016. Table.~\ref{tab:1} presents the average precision, recall, and ASR of each YOLOv5 variant under the attack of adversarial and random patches (IoU=0.5). The optimized adversarial patches significantly compromise detection performance. For example, YOLOv5-S achieves an ASR of 49.3\% under adversarial attack—far higher than the 10.5\% from random patches—demonstrating that the decline is not merely due to occlusion, but to the adversarial nature of the patch.
Furthermore, our method reduces YOLOv5-N’s precision from 99\% to 30.9\% and recall to 42.7\%, achieving 58\% ASR. These results confirm that the main factor driving performance degradation is the targeted adversarial design, not simple visual obstructions.

\begin{table}[htbp]
\centering
\caption{Comparison experiment results on YOLOv5 variants using the HRSC2016 dataset.}
\label{tab:1}
\resizebox{\columnwidth}{!}{
\begin{tabular}{llccc}
\toprule
\textbf{Model} & \textbf{Patch Type} & \textbf{Average Precision (\%)} & \textbf{Recall (\%)} & \textbf{ASR (\%)} \\
\midrule
\multirow{3}{*}{YOLOv5-M} 
    & No Patch           & 99.0 & 99.0 & 0.0 \\
    & Random Patch       & 92.7 & 93.3 & 7.1 \\
    & Adversarial Patch  & 38.5 & 42.7 & 54.1 \\
\midrule
\multirow{3}{*}{YOLOv5-S} 
    & No Patch           & 99.0 & 99.0 & 0.0 \\
    & Random Patch       & 88.3 & 89.7 & 10.5 \\
    & Adversarial Patch  & 38.5 & 43.7 & 49.3 \\
\midrule
\multirow{3}{*}{YOLOv5-N} 
    & No Patch           & 99.0 & 99.0 & 0.0 \\
    & Random Patch       & 88.8 & 89.9 & 7.6 \\
    & Adversarial Patch  & 30.9 & 35.7 & 58.0 \\
\bottomrule
\end{tabular}
}
\end{table}

\subsection{Ablation study}
This paper proposes to applies transformations locally to target regions instead of the global whole image, to reduce background interference and improve attack performance. To evaluate the effectiveness of the proposed strategy, additional comparison experiments are conducted using YOLOv5-M, YOLOv5-S, and YOLOv5-N: the baseline (no augmentation), global augmentation, and local augmentation. Table~\ref{tab:2} shows the comparison results.

Compared to the baseline, global augmentation led to mixed results: it improves performance on YOLOv5-M (recall drops by 11.2\%, ASR increases by 7.3\%), but degrads performance on YOLOv5-S and YOLOv5-N, with ASR on YOLOv5-S even decreasing by 2.5\%. In contrast, the local augmentation method consistently outperforms the others. On YOLOv5-M, it reduces recall by 16.6\% and improved ASR by 12.1\%. On YOLOv5-S, recall drops by 3.5\% while ASR increased by 1.1\%, again surpassing the global method.

For the lightweight YOLOv5-N, neither augmentation method brought clear benefits. Local augmentation performed similarly to the baseline (recall 35.7\% vs. 35.6\%, ASR 58.0\% vs. 58.8\%), and global augmentation slightly improved recall (38.0\%) but lowered ASR (56.3\%). This suggests that smaller models may lack sufficient feature extraction capacity, limiting the impact of fine-grained local perturbations.

Overall, the results, particularly the improvements on YOLOv5-M, demonstrate that the proposed local augmentation method effectively enhances adversarial patch robustness, especially for medium and large-scale detection models.

\begin{table}[htbp]
\centering
\small
\caption{Ablation study results on YOLOv5 variants on the HRSC2016 dataset}
\label{tab:2}
\begin{tabularx}{\columnwidth}{l l c c}
\toprule
\textbf{Model} & \textbf{Method} & \textbf{Recall} & \textbf{Attack Success Rate} \\
\midrule
\multirow{3}{*}{YOLOv5-M}
 & Baseline    & 59.3\% & 42.0\% \\
 & Global Aug  & 48.1\% & 49.3\% \\
 & Local Aug   & 42.7\% & 54.1\% \\
\midrule
\multirow{3}{*}{YOLOv5-S}
 & Baseline    & 47.2\% & 48.2\% \\
 & Global Aug  & 48.8\% & 45.7\% \\
 & Local Aug   & 43.7\% & 49.3\% \\
\midrule
\multirow{3}{*}{YOLOv5-N}
 & Baseline    & 35.6\% & 58.8\% \\
 & Global Aug  & 38.0\% & 56.3\% \\
 & Local Aug   & 35.7\% & 58.0\% \\
\bottomrule
\end{tabularx}
\end{table}

\subsection{Transferability Study}
To evaluate the transferability of adversarial patches, we conducted a series of experiments in which patches trained on three white-box models were tested on two black-box models. Table III compares the performance of the baseline, global augmentation, and local augmentation strategies across different YOLOv5 architectures.

The results indicate that while global augmentation can enhance transferability in certain cases, it is generally less effective than local augmentation. This is primarily due to the additional background noise introduced during global augmentation, which interferes with the optimization of adversarial patches.
In contrast, local augmentation significantly improves the generalization capability of the patches in most scenarios. For example, when YOLOv5-M is used as the white-box model, the attack success rates on YOLOv5-S and YOLOv5-N increase by 4.8\% and 5.4\%, respectively, compared to only 3.3\% and 3.9\% improvements achieved by global augmentation. Similarly, when YOLOv5-S serves as the white-box model, local augmentation results in improvements of 2.5\% and 2.1\%, whereas global augmentation leads to only a 1.1\% gain and even a -2.2\% drop, respectively.
These trends, however, are less evident when YOLOv5-N—a smaller model—is used as the white-box model. This may be attributed to the model’s limited feature extraction capacity, which restricts the effectiveness of fine-grained perturbations and diminishes the overall impact of both augmentation strategies. Despite this, local augmentation still performs better than global augmentation, causing only a 1.6\% drop on YOLOv5-M and achieving a 1.3\% increase on YOLOv5-S. In contrast, global augmentation results in a more substantial performance drop of 5.4\% on YOLOv5-M and 1.4\% on YOLOv5-S.

\begin{table}[htbp]
\centering
\caption{Transferability experiment results of attacks across different YOLOv5 architectures. }
\label{tab:3}
\resizebox{\columnwidth}{!}{ % 适配单栏宽度
\begin{tabular}{llccc}
\toprule
\textbf{Source Model} & \textbf{Method} & \textbf{YOLOv5-M} & \textbf{YOLOv5-S} & \textbf{YOLOv5-N} \\
\midrule
\multirow{3}{*}{YOLOv5-M} 
    & Baseline    & 42.0\% & 33.5\% & 34.3\% \\
    & Global Aug & 49.3\% & 36.8\% & 38.2\% \\
    & Local Aug  & 54.1\% & 38.3\% & 49.9\% \\
\midrule
\multirow{3}{*}{YOLOv5-S} 
    & Baseline    & 32.0\% & 48.2\% & 47.4\% \\
    & Global Aug & 33.1\% & 45.7\% & 45.2\% \\
    & Local Aug  & 34.5\% & 49.3\% & 49.5\% \\
\midrule
\multirow{3}{*}{YOLOv5-N} 
    & Baseline    & 32.9\% & 40.6\% & 58.8\% \\
    & Global Aug & 27.0\% & 39.2\% & 56.3\% \\
    & Local Aug  & 31.3\% & 41.9\% & 58.0\% \\
\bottomrule
\end{tabular}
}
\end{table}

\section{Conclusion}
To generate physically adversarial patches with higher transferability, this paper proposes a local augmentation based adversarial patch generation method, aiming to improve both the attack success rate and transferability of physical attacks in ship detection tasks. By applying local augmentation to only the target region instead of the entire image, interference to the background and irrelevant areas is reduced, thereby improving the stability of the patch optimization process. Experiments based on the HRSC2016 dataset demonstrate that the proposed method can increase the attack success rate of adversarial patches and enhance their transferability across different object detection models.

Although the proposed method effectively reduces the accuracy of ship detection models, its attack success rate remains relatively low compared to the object detection from natural images. In real-world scenarios, ships are typically large in size, and the adversarial patch is relatively small in comparison, making it difficult to cover the ship’s key features. As a result, the interference effect of the adversarial patches is relatively weaker. How to find the better  position to apply the adversarial patches will be the focus of our future work. Moreover, this study has not yet conducted physical-world tests of the adversarial patch. Future research will extend to real-world experiments to explore the impact of printed patches on actual systems.

\bibliographystyle{unsrt}
\bibliography{name} 
\end{document}